\documentclass[conference]{IEEEtran}
\IEEEoverridecommandlockouts

\usepackage{amsmath}
\usepackage{amssymb}
\usepackage{graphicx} 
\usepackage[numbers]{natbib}
\usepackage{setspace}
\usepackage{enumitem}

\usepackage{caption} 
\usepackage{mathtools}
\usepackage{balance}
\usepackage[dvipsnames,svgnames]{xcolor}
\usepackage{pifont}
\usepackage{array}
\usepackage{multirow}
 \usepackage[linesnumbered,ruled,vlined]{algorithm2e}
\usepackage{stackengine}
\usepackage{makecell}
\usepackage{multicol}
\usepackage{multirow}
\usepackage{caption}
\usepackage{subcaption}
\makeatletter
\newcommand{\thickhline}{%
    \noalign {\ifnum 0=`}\fi \hrule height 1pt
    \futurelet \reserved@a \@xhline
}
\newcolumntype{"}{@{\hskip\tabcolsep\vrule width 1pt\hskip\tabcolsep}}
\makeatother
\def\BibTeX{{\rm B\kern-.05em{\sc i\kern-.025em b}\kern-.08em
    T\kern-.1667em\lower.7ex\hbox{E}\kern-.125emX}}
\usepackage[colorlinks=true,urlcolor=Blue!90,linkcolor=Blue, citecolor=Blue]{hyperref}

\makeatletter

\begin{document}

\title{Enhancing the Performance of Automated Grade Prediction in MOOC using  Graph Representation Learning
\thanks{$^\dagger$These authors have an affiliation with \href{https://dsa.cs.usu.edu/}{Data Science and Applications Lab} at Utah State University.}}

    \makeatletter
\newcommand{\linebreakand}{%
  \end{@IEEEauthorhalign}
  \hfill\mbox{}\par
  \mbox{}\hfill\begin{@IEEEauthorhalign}
}
\makeatother

\author{\IEEEauthorblockN{Soheila Farokhi$^\dagger$}
\IEEEauthorblockA{
Utah State University\\
\href{mailto:soheila.farokhi@usu.edu}{soheila.farokhi@usu.edu} }
\and
\IEEEauthorblockN{Aswani Yaramal$^\dagger$}
\IEEEauthorblockA{
Utah State University\\
\href{mailto:aswani.yaramala@usu.edu}{aswani.yaramala@usu.edu} }
\and
\IEEEauthorblockN{Jiangtao	Huang}
\IEEEauthorblockA{  
Nanning Normal University\\
\href{mailto:jiangtao@nnnu.edu.cn}{jiangtao@nnnu.edu.cn} }\and
\IEEEauthorblockN{Muhammad Fawad Akbar Khan$^\dagger$} 
\IEEEauthorblockA{
Utah State University\\
\href{mailto:khan@usu.edu}{khan@usu.edu} }
\linebreakand \IEEEauthorblockN{Xiaojun Qi}
\IEEEauthorblockA{ 
Utah State University\\
\href{mailto:xiaojun.qi@usu.edu}{xiaojun.qi@usu.edu}
}\and
\IEEEauthorblockN{Hamid Karimi$^\dagger$}
\IEEEauthorblockA{
Utah State University\\
\href{mailto:hamid.karimi@usu.edu}{hamid.karimi@usu.edu}
}

}

\maketitle
\begin{abstract}

In recent years, Massive Open Online Courses (MOOCs) have gained significant traction as a rapidly growing phenomenon in online learning. Unlike traditional classrooms, MOOCs offer a unique opportunity to cater to a diverse audience from different backgrounds and geographical locations. Renowned universities and MOOC-specific providers, such as Coursera, offer MOOC courses on various subjects. Automated assessment tasks like grade and early dropout predictions are necessary due to the high enrollment and limited direct interaction between teachers and learners. However, current automated assessment approaches overlook the structural links between different entities involved in the downstream tasks, such as the students and courses. Our hypothesis suggests that these structural relationships, manifested through an interaction graph, contain valuable information that can enhance the performance of the task at hand. To validate this, we  construct a unique knowledge graph for a large MOOC dataset, which will be publicly available to the research community.
Furthermore, we utilize graph embedding techniques to extract latent structural information encoded in the interactions between entities in the dataset. These techniques do not require ground truth labels and can be utilized for various tasks. Finally, by combining entity-specific features, behavioral features, and extracted structural features, we  enhance the performance of predictive machine learning models in student assignment grade prediction. Our experiments demonstrate that structural features can significantly improve the predictive performance of downstream assessment tasks. The code and data are available in \url{https://github.com/DSAatUSU/MOOPer_grade_prediction}

\end{abstract}


\section{Introduction}
\label{sec:introduction}

Massive Open Online Courses (MOOCs) are open courses with open enrollment that aim to make educational material accessible to a large audience online and often for free regardless of location or background~\cite{christensen2013mooc}.
Over the last ten years, MOOCs have emerged as a new online learning trend. In addition to dedicated providers (e.g., Coursera and Udemy), many renowned universities have launched numerous courses that have drawn an oversized enrollment. MOOCs play a crucial role in modern education by democratizing access to quality learning~\cite{dillahunt2014democratizing}, promoting lifelong learning~\cite{ossiannilsson2021moocs}, and offering flexibility~\cite{papadakis2023moocs}, cost-effectiveness~\cite{castillo2015moocs}, and customization~\cite{daradoumis2013review}. In addition, they facilitate global collaboration, rapid dissemination of knowledge~\cite{mcauley2010mooc}, and targeted skill development~\cite{czerniewicz2014developing} while providing a platform for educators to experiment with innovative teaching methods. By expanding the reach of educational institutions beyond traditional borders, MOOCs foster a global community of learners and educators, ultimately transforming the education landscape and making learning more accessible, flexible, and personalized. Automated grade prediction in MOOCs is a vital task with far-reaching benefits for students, instructors, and course providers. Some of these benefits include:

    \begin{itemize}[leftmargin=0.5cm]
        \item [\ding{114}] \textbf{Personalized learning:} Automated grade prediction can help identify a student's strengths and weaknesses, enabling the creation of personalized learning paths to address specific knowledge gaps~\cite{qi2018temporal,fan2022interpretable}.
   
        \item [\ding{114}]  \textbf{Early intervention:} By predicting a student's performance, instructors and course providers can intervene early and offer additional support or resources to help struggling students succeed~\cite{yang2017behavior,adnan2021predicting}.
 
        \item [\ding{114}] \textbf{Improved engagement:} With insights into student performance, course providers can make data-driven decisions to improve course design, engagement, and retention~\cite{lemay2020grade}.
     
        \item [\ding{114}] \textbf{Enhanced feedback:} Automated prediction systems can provide immediate feedback to students, which can help them understand their progress and make necessary adjustments in their learning strategies~\cite{lemay2020grade}.

        \item [\ding{114}] \textbf{Motivation:}   Providing students with their predicted grades can encourage them to take control of their learning and motivate them to improve their performance~\cite{de2016role,xu2016motivation}.
     
        \item [\ding{114}]  \textbf{Performance tracking:} Automated grade prediction enables tracking of student performance over time~\cite{ren2016predicting}, helping instructors identify trends and patterns that can inform future course design and delivery.
       
        \item [\ding{114}]  \textbf{Dropout prevention:} Identifying students at risk of dropping out or under-performing early on, allows for targeted interventions to help retain students and improve overall course completion rates~\cite{chen2017mooc,xing2019dropout}.
    
        \item [\ding{114}] \textbf{Informed decision-making:} Grade prediction can help course providers evaluate the effectiveness of their courses, identify areas for improvement, and make evidence-based decisions about future course offerings~\cite{tzeng2022mooc}.
      
        \item [\ding{114}] \textbf{Scalability:} Automated grade prediction can be especially beneficial in MOOCs, which often have large numbers of students~\cite{chauhan2014massive} because it can provide a more consistent and scalable way to predict and monitor student performance across diverse groups.
    \end{itemize}

\begin{figure}[h]
\centering
\includegraphics[width=\columnwidth]{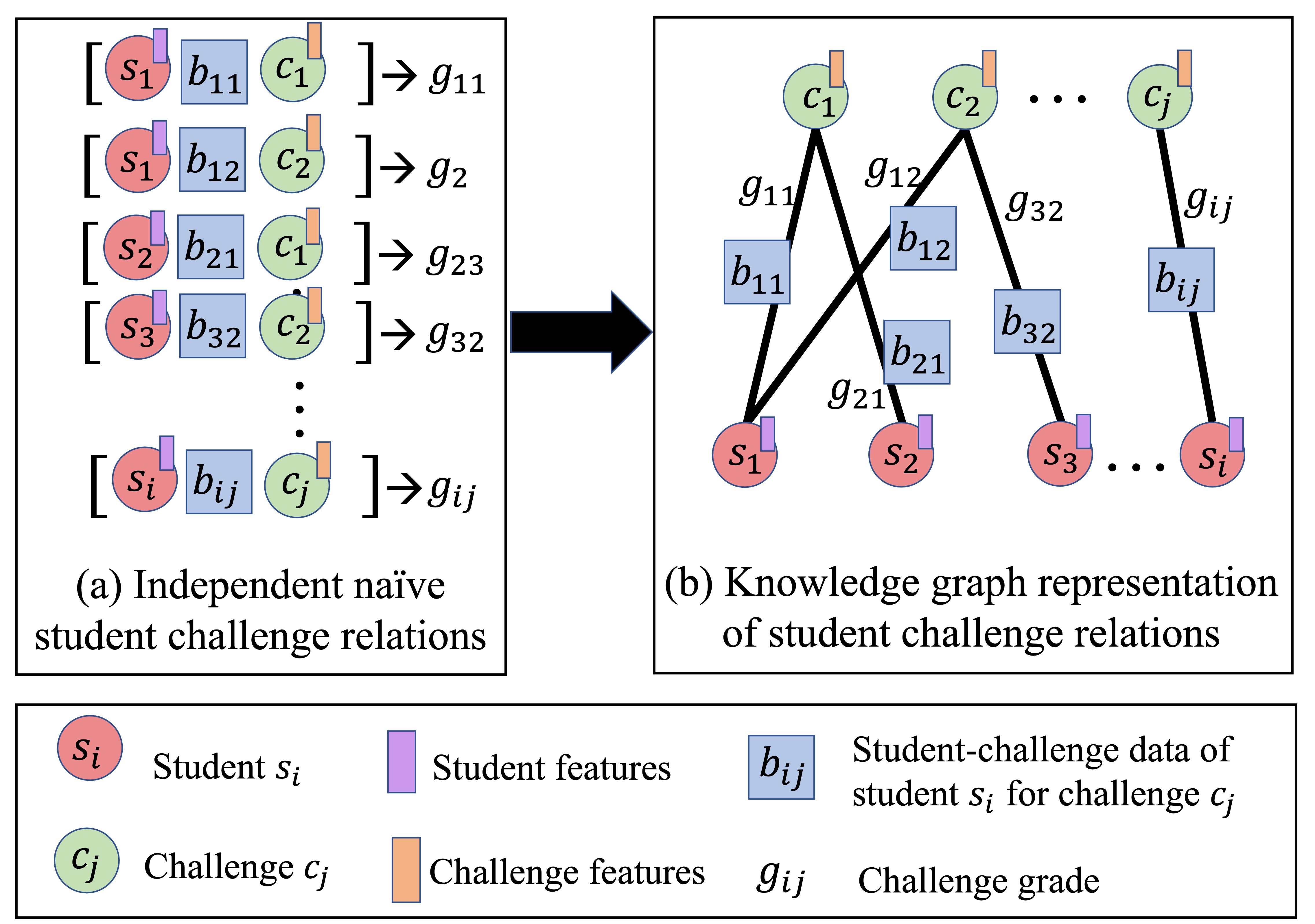}

\caption{Visualizing the traditional approach used in prior prediction models compared to our graph representation.}
\label{fig:knowledge graph}

\end{figure}

Over the past decade, numerous automated grade prediction methods have emerged in the MOOC landscape, offering various benefits, as mentioned above. However, existing methods still face two critical shortcomings. Firstly, current methods have a large granularity of grade prediction tasks, often focusing on significant assignments or entire courses~\cite{moreno2018prediction}. In contrast, MOOC systems frequently include shorter exercises that target specific skills, such as Python's `for loop' and `continue statement.' Predicting future grades for these smaller exercises would be highly beneficial. In this paper, we refer to such exercises as `\texttt{challenges}'--See Section~\ref{sec:Dataset}. Secondly, most existing methods overlook the complex yet rich structural relationships between pertinent entities, such as students and courses, which are essential to the prediction task. To illustrate, consider the visualization in Figure~\ref{fig:knowledge graph}, which demonstrates our \texttt{challenge} grade prediction study. Figure~\ref{fig:knowledge graph}(a) shows a scenario where the data of each student-\texttt{challenge} pair is treated separately, whereas Figure~\ref{fig:knowledge graph}(b) presents our approach, which conceptualizes the relationship between relevant entities as an interaction graph. This graph encodes the inherent links between students and \texttt{challenges}. We hypothesize that the interaction graph between entities offers rich structural information, which can be extracted and leveraged to enhance the predictive performance of the grade prediction model.

In this study, we address the above-mentioned shortcomings by focusing on the \texttt{challenge} prediction task, which offers higher granularity than course outcome prediction. First, we introduce a new dataset from a large MOOC provider in China, containing various entities (e.g., \texttt{challenge} , course, chapter) for thousands of students. We then construct an interaction graph between students and \texttt{challenges} as a bipartite graph, as seen in Figure~\ref{fig:knowledge graph}(b), and extract salient and dense entity-level vector representations using advanced graph representation learning techniques. Specifically, we employ two powerful node embedding learning methods, node2vec~\cite{grover2016node2vec} and DeepWalk~\cite{perozzi2014deepwalk}. One of the main benefits of these graph representation learning methods is their unsupervised training, using only the underlying graph structure without ground truth labels, such as \texttt{challenge} grades. To validate the effectiveness of knowledge graph representations, we develop machine learning prediction models and conduct extensive experiments with our dataset. Our results show that incorporating structural graph information can enhance the predictive performance of \texttt{challenge} grade prediction. Worth noting that existing methods often rely on different and often complicated data and features, such as student forums~\cite{xu2016motivation} or watched  videos~\cite{yang2017behavior}, which may only sometimes be readily available.

    \begin{itemize}
        \item [\ding{228}] We introduce a new MOOC dataset that includes various entities and their interactions for several thousand students, which will be publicly available to the research community.
        
      \item [\ding{228}]  Contrary to current grade prediction methods in MOOCs, we focus on short, small, and particular exercises, referred to as \texttt{challenges} in our dataset.

        \item [\ding{228}]  We build a bipartite graph between students and \texttt{challenges}, then extract unsupervised dense entity representations using advanced neural networks.

        \item [\ding{228}]  We conduct extensive experiments, demonstrating the usability of the interaction graph and its resulting representations for \texttt{challenge} grade prediction.
    \end{itemize}

\noindent \textbf{Remark.} In this paper, the term ``performance" is used to describe two different concepts. First, it refers to the academic achievement of students, specifically their grades. Second, it pertains to the performance of machine learning algorithms, specifically their accuracy. The distinction between these two uses of ``performance" is made clear from the context in which it is used.\\     
The rest of this paper is organized as follows. First, in Section~\ref{sec:related}, we briefly review related work. Next, in Section~\ref{sec:Dataset}, we introduce the dataset, followed by our grade prediction problem statement in Section~\ref{sec:problem}. Section~\ref{sec:method} describes the methodology, and Section~\ref{sec:experiments} includes experiments and discussions. Finally, we conclude the paper in Section~\ref{sec:conclusion} and enumerate several possible future directions.

\section{Related Work}
\label{sec:related}

Educational data mining is an emerging field that leverages computational approaches to analyze large-scale educational data~\cite{karimi2020deep,karimi2020online,karimi2019roadmap,karimi2020towards,karimi2022teachers,knake2021educational,karimi2021automatic}. In particular, computational techniques for performance prediction in MOOCs were developed over the past ten years as MOOCs gained popularity. In~\cite{ren2016predicting}, the authors developed a personalized linear multiple regression model to predict a student's future performance for specific assessment activities within a MOOC. They extracted six types of features (session features, quiz-related features, video-related features, homework-related features, time-related features, and interval-based features) from MOOC server logs to identify learning behavior and study habits for different students. Ramesh et al.~\cite{ramesh2013modeling} developed two distinct probabilistic soft logic models to forecast student achievement. One model used behavioral, linguistic, and structural features, while the other treated learner engagement as a latent variable and linked observed features to one or more types of engagement. In ~\cite{xu2016motivation}, the authors first classified learners according to motivation into three groups: \emph{certification earning}, \emph{video watching}, and \emph{course sampling}. They then used an SVM-based model to predict grades by classifying certification learners into two classes (may or may not obtain a certification). Brinton and Chiang~\cite{brinton2015mooc} applied standard algorithms to predict whether a MOOC student would be Correct on First Attempt (CFA) in answering a question. They found that parsing click-stream data into summary quantities was useful for classifying CFA. They computed nine video-watching summary quantities, such as fraction spent, fraction completed, fraction played, number of pauses, fraction paused, average playback rate, standard deviation of playback rate, number of rewinds, and number of fast forwards. Robust CFA prediction provided insights into student learning and helped instructors address students collectively. A continuous assessment prediction problem could be viewed as knowledge tracing, which predicted student's future performance based on their past activity. In~\cite{yang2017behavior}, a time series neural network was trained using lecture video watching click-streams for predicting student grades in MOOCs. Piech et al.~\cite{piech2015deep} applied RNN and LSTM networks to predict students' responses to exercises based on their past activity in MOOCs.
Furthermore, \cite{qi2018temporal} used LSTM, RNN, LR, and SVM to make individual grade predictions, with LSTM achieving the best performance with an average AUC score of 0.748. Some researchers proposed novel deep learning models to improve MOOC learner performance prediction.
In~\cite{qi2018temporal}, the authors developed a unified model that incorporated student demographics, forum activities, and learning behavior to predict students' assignment grades. They cast the prediction task as a binary classification problem, predicting whether the student's grade was ranked in the top 30\% of all students.
Kim et al.~\cite{kim2018gritnet} used deep learning methods to predict real-time student performance. They proposed a new deep learning-based method (GritNet), built upon the bidirectional long short-term memory (BLSTM) to predict student performance, such as graduation prediction. The algorithm significantly improved prediction quality within the first few weeks of the student experience. In~\cite{karimi2020online}, the authors introduced Deep Online Performance Evaluation (DOPE) to predict a student's performance in a specific course. Their method first represented the online learning system as a knowledge graph and learned student and course embeddings from historical data using a relational graph neural network. Additionally, DOPE used an LSTM to harness student behavior data into a condensed encoding since the data naturally took on a sequential shape. In~\cite{li2020peer}, the authors introduced a model using Graph Neural Networks called \begin{math}R^2GCN\end{math} (GNNs). They first created a network of questions and interactions among students, and then they utilized the proposed model to predict students' performance. Specifically, they formulated the student performance prediction in interactive online question pools as a node classification problem on a large heterogeneous network consisting of questions, students, and the interactions between them, better capturing the underlying relationship between questions and students.

All the previous studies on grade prediction in MOOCs mainly focus on course-level outcomes or major assignments, but little attention is paid to predicting grades for short, small, and specific exercises, which are called \texttt{challenges} in this study. Additionally, most existing methods ignore the complex structural relationships between pertinent entities, such as students and courses. To address these limitations, this study proposes a new approach that constructs an interaction graph between students and \texttt{challenges} and leverages advanced graph representation learning techniques to extract salient and dense entity-level vector representations. The effectiveness of this approach is validated through machine learning prediction models and extensive experiments on a new MOOC dataset introduced in this study.
\section{Dataset}
\label{sec:Dataset}

MOOPer is an extensive online open practice dataset provided by a prominent Chinese university. The dataset contains students' online practice data from 2018 to 2019, along with supplementary information on users and practice projects organized in the form of knowledge maps. The MOOPer dataset is divided into two components: interactive data and knowledge maps. The interactive dataset comprises 2.53 million practice records, 21.6 million system feedback records, and 15,000 forum chat records. In the knowledge map section, there are 11 unique entity categories and ten specific relationship types.
\subsection{Knowledge map}
The knowledge map provides information about entities and their relations. The structure of the knowledge map is illustrated in the left section of Figure~\ref{fig:entities and interactions}.
\begin{figure}[ht]
\centering \includegraphics[width=\columnwidth]{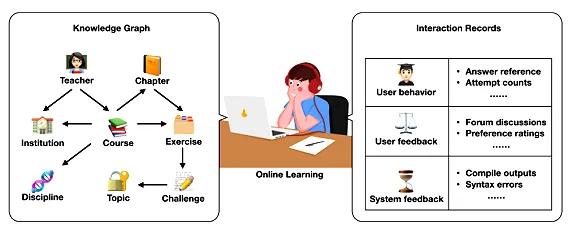}
\caption{Entities and their relations in the MOOPer dataset.}
\label{fig:entities and interactions}
\end{figure}

\subsubsection{Entities} The dataset includes entities such as discipline, sub-discipline, department, courses, chapter, \texttt{challenge}, exercise, school, student, teacher, and topic. More details about each entity will be provided in the dataset description upon its release.

\subsubsection{Relation} This section explains the relationships between different entities. There is a hierarchical relationship among the entities. For instance, each course contains numerous exercises, and each exercise comprises multiple challenges. However, a single exercise can be linked to multiple courses.

\subsection{Interactive data}
Student engagement with learning materials is categorized into three groups: user behavior, user feedback, and system feedback, described below.

\subsubsection{User behavior}
This category includes information on the \texttt{challenges} students completed and the grades they achieved for those tasks. It also contains data on the interaction's open and close time, the number of attempts to solve the challenge, whether students referred to the answers, and other additional information.

\subsubsection{User feedback}
This category displays the user's rating of the interactive practice activities and their choice of learning resources with different difficulty levels and problem types. Moreover, students can share their discussions in the forum. Chat content can be used to assess students' learning states and satisfaction, while the question-and-answer conversation can reveal their "blind spots" in knowledge acquisition. Additionally, the user's behavior in the forum serves as a critical indicator to speculate on their psychological condition and learning style~\cite{fang2019social}.

\subsubsection{System feedback}
This category provides students with feedback on the system's results, such as syntax errors in their code, compilation results of the submitted code, and the discrepancy between the actual output and the intended output. Such data can be used to better understand users' learning abilities and knowledge mastery~\cite{shatnawi2014automatic}.
\section{Problem Statement}
\label{sec:problem}
Suppose we have a subset of $m$ \texttt{challenges} from our dataset denoted as $\mathcal{C} = \{c_1, c_2, \cdots, c_m\}$. Also, let there be $n$ students who have interacted with (or attempted) at least two of the $m$ \texttt{challenges} in $\mathcal{C}$, denoted as $\mathcal{S} = \{s_1, s_2, \cdots, s_n\}$. For each \texttt{challenge} $c_j$, we assume the features can be represented as the vector ${\bf f}_j \in \mathbb{R}^{d_c}$, where $d_c$ is the dimension size after encoding the \texttt{challenge} features. All \texttt{challenge} features are denoted as $\mathcal{F} \in \mathbb{R}^{d_c \times |\mathcal{C}|}$. Similarly, for each student $s_i$, we assume the collected information can be represented as the vector ${\bf d}_i$ $\in \mathbb{R}^{d_s}$, with $d_s$ being the dimension size after encoding the student-related data. All student features are denoted as $\mathcal{D} \in \mathbb{R}^{d_s \times |\mathcal{S}|}$. In addition to the student-specific and \texttt{challenge}-specific data, the system is assumed to have collected some student-\texttt{challenge} data for each student $s_i$ engaging with \texttt{challenge} $c_j$, which we represent as ${\bf b}_{ij}$. Let ${\it g}_{ij}$ denote the grade of student $s_i$ in \texttt{challenge} $c_j$. Let $\mathcal{B}$ and $\mathcal{G}$ represent the entire student-\texttt{challenge} data and grades, respectively. Moreover, suppose graph $\mathrm{BG}=(\mathcal{S}, \mathcal{C},E)$ denotes the bipartite graph of student-\texttt{challenge} interactions, where $E \subset \mathcal{S} \times \mathcal{C}$ and $e=(s_i, c_j) \in E $ is the edge between student $s_i$ and \texttt{challenge} $c_j$. Finally, let $\mathcal{M}(.)$ denote the graph representation machinery that takes graph $\mathrm{BG}$ and provides dense representations for both student and \texttt{challenge} pairs $s_i$ and $c_j$ in $E$.

Given the notations listed above, our goal is to learn a machine learning model $f(.)$ that can predict students' \texttt{challenge} outcomes $\mathcal{G}$ as follows:
\vspace{-0.25ex}
\begin{align}
f((\mathcal{C}, \mathcal{S}, \mathcal{F}, \mathcal{D}, \mathcal{B}, \mathcal{M}(\mathrm{BG})), \mathcal{G}) \rightarrow \hat{\mathcal{G}} \nonumber
\end{align}

\noindent where $\hat{\mathcal{G}}$ should be as close as possible to $\mathcal{G}$.
\section{Methodology}
\label{sec:method}

Our methodology comprises several components. First, in Section~\ref{sec:mooc-features}, we describe extracted MOOC-related features. Next, in Section~\ref{sec:structural}, we detail the structural features from constructed interaction graph. Finally, in Section~\ref{sec:models}, we explain the developed machine learning models for grade prediction. 

\subsection{MOOC-related Features}
\label{sec:mooc-features}
First, we extracted available MOOC-related features in Table~\ref{tab:mooper_features}. Some of these features are specific to a particular entity; for example, the difficulty is specific to the \texttt{challenge} entity. In addition, some features describe the interaction between the entities; for example, the final score is the student's score for a specific \texttt{challenge}.
\begin{table}[h]\small
\setlength{\tabcolsep}{3pt}
\def\arraystretch{1.2}
\caption{MOOC-related features}
    \centering
    \begin{tabular}{l|l} \hline
         \textbf{Feature} & \textbf{Description} \\\hline \hline
        User ID& Student unique identification number\\
        \texttt{Challenge} ID& \texttt{Challenge} unique identification number 
        \\
        Timestamp&The time when the student first opened the \texttt{challenge}\\
        Final Score& Final score of student on \texttt{challenge}\\
        Exercise ID&Exercise ID for a particular \texttt{challenge}\\
        Course ID& Course ID of a specific exercise\\
        Difficulty&Difficulty level of a \texttt{challenge}\\
        \#Retries&Number of attempts to complete a \texttt{challenge}\\
        Duration&Time spent to complete a \texttt{challenge}\\
        \hline

    \end{tabular}
    \label{tab:mooper_features}
\end{table}

\subsection{Structural Features}
\label{sec:structural}
 
As described in Section~\ref{sec:problem}, we constructed a bipartite graph by considering students and \texttt{challenges} as nodes and interactions between these nodes as edges. Table~\ref{tab:KGstatistics} shows some of the properties of this graph.

\begin{table}[h]\small
\centering
\caption{Properties of constructed interaction graph}
\label{tab:KGstatistics}
\begin{tabular}{l|c}
\hline
Type &	Bipartite\\\hline
Includes node features&	Yes\\\hline
Includes edge features&	Yes\\\hline
Number of students ($|\mathcal{S}|$) & 5537  \\\hline
Number of \texttt{challenges}  ($|\mathcal{C}|$) & 1981 \\\hline
Number of nodes (students+\texttt{challenges} or $|\mathcal{S}| + |\mathcal{C}|$) & 7518 \\\hline
Number of interactions (edges or $|E|$) & 115124 \\\hline
Density & 0.01 \\\hline
\end{tabular}

\end{table}

From graph $BG$, we first extracted two node-level pieces of information, including the degree and eigenvector centrality of the \texttt{challenges} and users. The degree of a node is determined by the number of edges incident to it. Eigenvector centrality measures a node's influence within a network~\cite{bonacich2007some}. Next, we extracted node embeddings for the users and \texttt{challenges} using two well-known graph representation learning methods, namely node2vec~\cite{grover2016node2vec} and DeepWalk~\cite{perozzi2014deepwalk}. Node2vec is an algorithmic framework for learning continuous feature representations for graph nodes. In node2vec, we learn a mapping of nodes to a low-dimensional space of features that maximizes the likelihood of preserving network neighborhoods of nodes~\cite{grover2016node2vec}. DeepWalk is a novel approach for learning latent representations of vertices in a network. Statistical models can utilize these latent representations to encode social relations in a continuous vector space~\cite{perozzi2014deepwalk}. During the experiments, we replaced the user ID and \texttt{challenge} ID in the dataset with these structural features when structural data were to be used. Table~\ref{tab:graph_features} shows the extracted structural features that we employed.

\begin{table}[h]\small
\setlength{\tabcolsep}{3pt}
\def\arraystretch{1.2}
\caption{Structural features}
    \centering
    \begin{tabular}{l|l} \hline
         \textbf{Feature} & \textbf{Description} \\\hline \hline
        \texttt{Challenge} embedding & \texttt{Challenge} node embedding\\
        User embedding& User node embedding\\
        Degree of \texttt{challenge}&Number of edges linked to a \texttt{challenge}\\
        Degree of user& Number of edges connected to user\\
        \texttt{Challenge} EC& Eigenvector centrality of \texttt{challenge}\\
        User EC& Eigenvector centrality of user \\ \hline
        
    \end{tabular}
    \label{tab:graph_features}
\end{table}

\vspace{-0.48cm}
\subsection{Prediction Models}
\label{sec:models}
We utilized the following prediction models to predict a student's grade in a \texttt{challenge}:

\begin{itemize}
    \item  [\ding{111}] Random Forest \\
Random Forest is a classifier that employs multiple decision trees on various subsets of the input dataset and averages the results to enhance the predicted accuracy. It is based on ensemble learning, which combines multiple classifiers to tackle a challenging problem and improve the model's performance. Formally, given a dataset $D = \{(x_i, y_i)\}_{i=1}^n$, Random Forest builds $T$ decision trees $\{h_t(x)\}_{t=1}^T$ using bootstrapped subsets of $D$. The final prediction is obtained by averaging the individual tree predictions:
\begin{equation}
    H(x) = \frac{1}{T} \sum_{t=1}^T h_t(x)
\end{equation}

    \item  [\ding{111}] Gradient Boosting \\
Gradient boosting is a machine learning technique for regression and classification problems that produces a prediction model as an ensemble of weak prediction models. Gradient boosting is an iterative process that effectively transforms a weak learner into a strong learner. Formally, given a dataset $D = \{(x_i, y_i)\}_{i=1}^n$, gradient boosting builds an ensemble of weak learners $\{h_t(x)\}_{t=1}^T$ by minimizing the loss function $L(y, F(x))$:
\begin{equation}
    F(x) = \sum_{t=1}^T \alpha_t h_t(x)
\end{equation}
where $\alpha_t$ is the step size at iteration $t$.

    \item  [\ding{111}] XGBoost \\
XGBoost is an implementation of gradient-boosted decision trees. In this approach, decision trees are generated sequentially. Weights play a crucial role in XGBoost. Each independent variable is weighted before being input into the decision tree that predicts outcomes. Formally, given a dataset $D = \{(x_i, y_i)\}_{i=1}^n$, XGBoost builds an ensemble of $T$ decision trees $\{h_t(x)\}_{t=1}^T$ by minimizing the regularized loss function $L(y, F(x)) + \Omega(h_t)$:
\begin{equation}
    F(x) = \sum_{t=1}^T h_t(x)
\end{equation}
where $\Omega(h_t)$ is the regularization term that controls the complexity of the decision trees.
\end{itemize}

\noindent\textbf{Remarks}. We performed experiments using alternative prediction models, such as Support Vector Machine (SVM) and Decision Tree Classifier (DT), but observed unsatisfactory performance with our data. As a result, we chose to utilize the three prediction models discussed in our analysis. Moreover, our objective is not to develop a novel machine learning methodology. Instead, we aim to examine the influence of structural features in a MOOC-induced graph. Thus, we employed off-the-shelf models. We plan to explore more advanced models in future work.

\section{Experiments}
\label{sec:experiments}

To evaluate the effectiveness of structural features in \texttt{challenge} grade prediction, we performed various experiments, which are discussed in this section. First, in Section~\ref{subsec:exp}, we describe the experimental settings. Next, in Section~\ref{subsec:results}, we present the main empirical results. Section~\ref{subsec:student_levels} includes experimental results showing the association between model predictions across  students' academic performance levels. Finally, in Section~\ref{subsec:features}, we analyze the importance of different features in grade prediction performance.

\subsection{Experimental Settings}
\label{subsec:exp}
We divided the final scores into five classes to perform grade classification. The grade range for each class and the number of instances with that grade class is displayed in Table~\ref{tab:cl_scores}.
 
\begin{table}[h]\small
\centering
\caption{\label{tab:cl_scores}Discretized grades and number of instances in each grade label}
\begin{tabular}{l|c|c}
\textbf{Class} & \textbf{Grade range} &\textbf{\# Instances}\\\hline \hline
class 0  & $0 \leq grade < 20$ & 34500\\
class 1   & $20 \leq grade < 40$ & 34403\\
class 2 & $40 \leq grade < 60$ & 19315\\
class 3 & $60 \leq grade < 80$&3970\\
class 4 & $80 \leq grade \leq 100$&34500\\
\end{tabular}

\end{table}

To split the dataset into train and test sets, for every student, we first sorted the \texttt{challenges} they have completed based on their timestamp. Then, we added approximately the first 80\% of \texttt{challenges} for each student in the training set and the rest in the test set. The reason for performing this type of split is to predict a student's future grades based on their previous performance. This way, the prediction reflects the practical usage of grade prediction tasks where we are interested in estimating the future grades during an academic period, for example, to intervene and offer help to possible struggling students.

The Random Forest and Gradient boosting methods were implemented using the scikit-learn package. To implement the XGBoost model, we used the XGBoost Python package. In addition, we utilized the node2vec package\footnote{\url{https://pypi.org/project/node2vec/}} and the Karateclub package\footnote{\url{https://karateclub.readthedocs.io/en/latest/}} for node2vec and Deepwalk implementation, respectively. The hyperparameters used in learning node embeddings with node2vec and DeepWalk are as follows: embedding dimension: 128, number of walks: 100, walk length: 10, and window size: 10.

\textbf{Evaluation Metrics.}
The performance prediction models were evaluated using a variety of measures. Suppose TP, FP, TN, and FN are the number of true positive, false positive, true negative, and false negative samples, respectively. Given this notation, we reported the prediction performance using the following metrics. 

\textit{Accuracy} calculates the proportion of accurate predictions to the total number of instances considered:

$$Accuracy = \dfrac{TP + TN}{TP+FP+TN+FN}$$

\textit{Precision} is the ratio of accurately predicted positive patterns in a positive class to all predicted positive patterns:

$$Precision = \dfrac{TP}{TP+FP}$$

\textit{Recall} measures the fraction of positive patterns that are correctly classified:

$$Recall = \dfrac{TP}{TP+FN}$$

\textit{F1-score} represents the harmonic mean of the Precision and Recall:
\vspace{-0.25cm}
$$F1\text{-}score = 2 \times \dfrac{Precision \times Recall}{Precision + Recall}$$

\textit{ROC (Receiver Operating Characteristic)} curve is a graph that displays how well a classification model performs across all classification thresholds. It is a probability curve that plots the TPR (True Positive Rate) and FPR (False Positive Rate) at different threshold values.

\textit{AUC} calculates the total two-dimensional region under the complete ROC curve.

\subsection{Experimental Results}
\label{subsec:results}

\begin{table}[h]\small
\renewrobustcmd{\bfseries}{\fontseries{b}\selectfont}
\renewrobustcmd{\boldmath}{}
\newrobustcmd{\B}{\bfseries}
\setlength{\tabcolsep}{2pt}
\centering
\vspace{-0.2cm}
\caption{\label{cl_rf_cr}Comparing the performance of 9 models on grade prediction task. The best algorithm in each column is displayed in bold.}
\vspace{-0.3cm}
\begin{tabular}{c|cccc}
Model & Accuracy & Precision & Recall & F1-score\\\hline
Random Forest   & 0.80 &    0.85 &     0.86&      0.86  \\
XGBoost    & 0.78 &  0.84  &    0.85 &     0.85  \\
Gradient Boosting    &  0.84 &   0.89   &   0.89  &    0.89  \\\hline
Random Forest + DeepWalk   & 0.89 &    0.92 &     0.92&      0.92  \\
XGBoost + DeepWalk   & 0.87 &  0.91  &    0.90 &     0.90  \\
Gradient Boosting + DeepWalk    & 0.89 &  0.92   &   0.92  &   0.92  \\\hline
Random Forest + node2vec      & 0.88 &  0.92    &  0.91   &   0.91   \\
XGBoost + node2vec      & 0.89 & 0.92     & 0.92    &  0.92   \\
Gradient Boosting + node2vec   & \B 0.89 & \B 0.92   &  \B 0.92  &  \B  0.92  \\
\end{tabular}
\end{table}
\begin{figure*}[htbp]
\centering
\noindent\begin{minipage}[b]{.33\textwidth}
\centering
\includegraphics[width=0.85\linewidth]{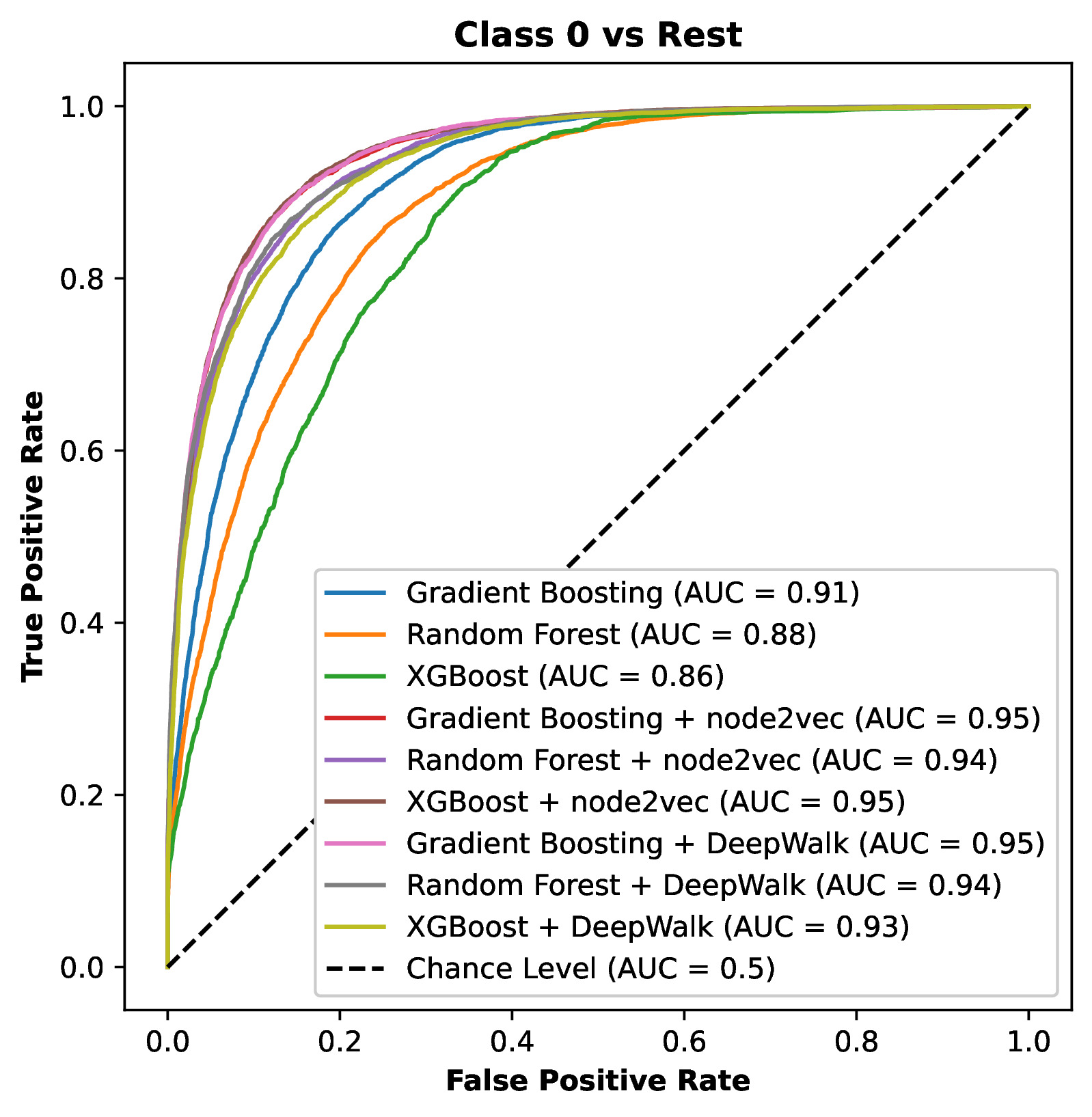}
\caption*{}
\end{minipage}\hfill
\begin{minipage}[b]{.33\textwidth}
\centering
\includegraphics[width=0.85\linewidth]{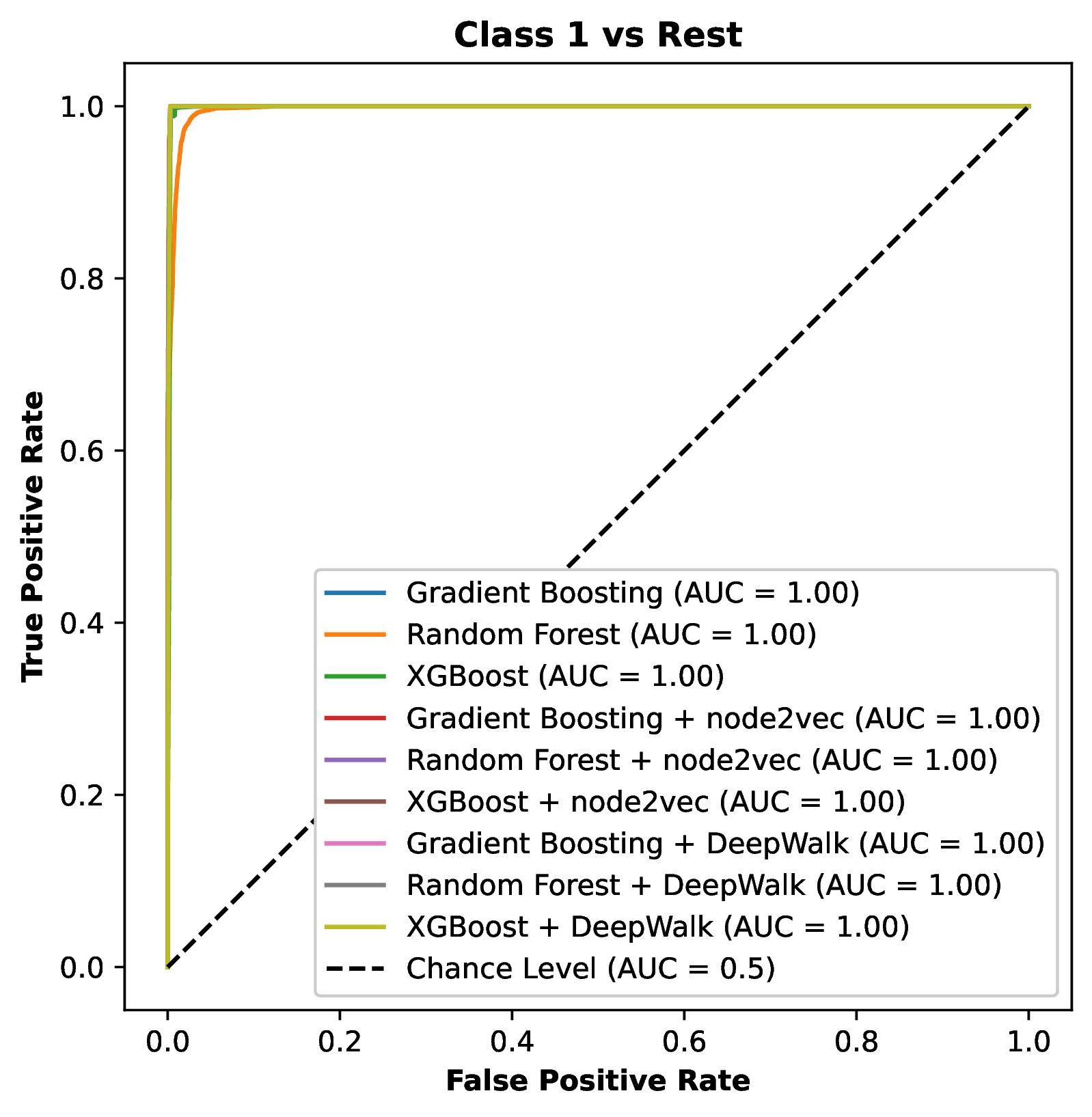}
\caption*{}
\end{minipage}\hfill
\begin{minipage}[b]{.33\textwidth}
\centering
\includegraphics[width=0.85\linewidth]{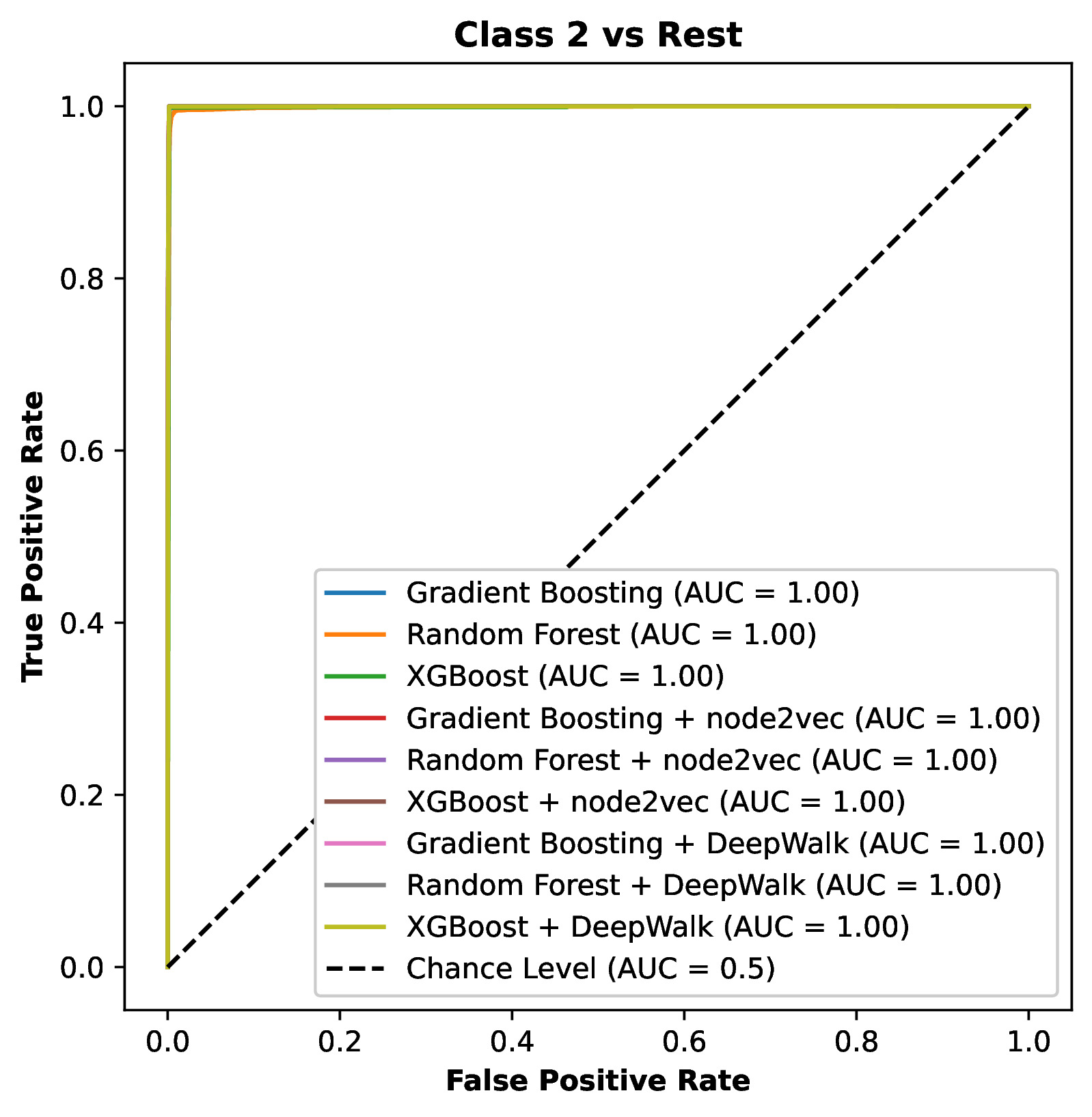}
\caption*{}
\end{minipage}\hfill
\begin{minipage}[b]{.5\textwidth}
\centering
\includegraphics[width=0.55\linewidth]{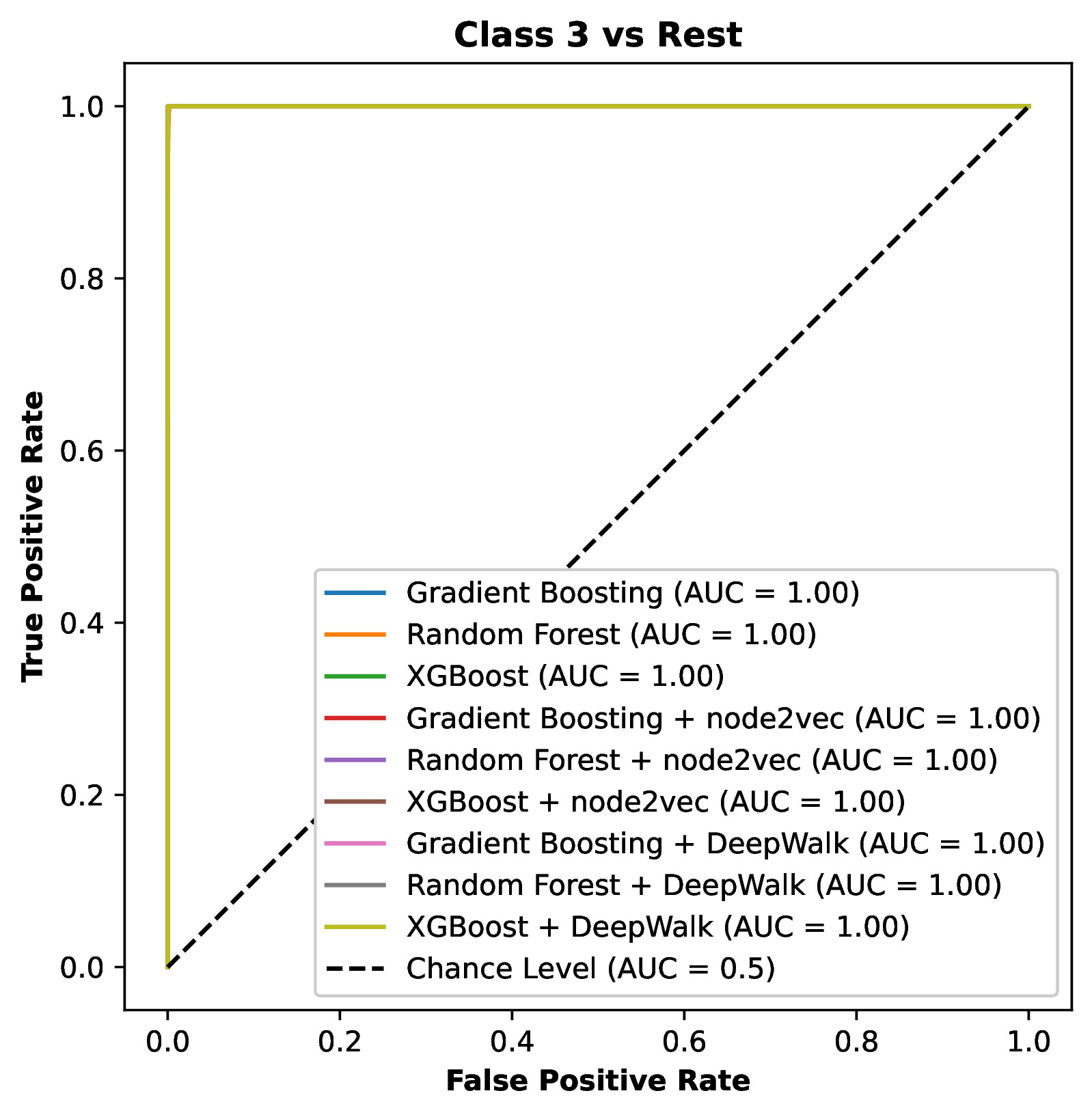}
\caption*{}
\end{minipage}\hfill
\begin{minipage}[b]{.5\textwidth}
\centering
\includegraphics[width=0.55\linewidth]{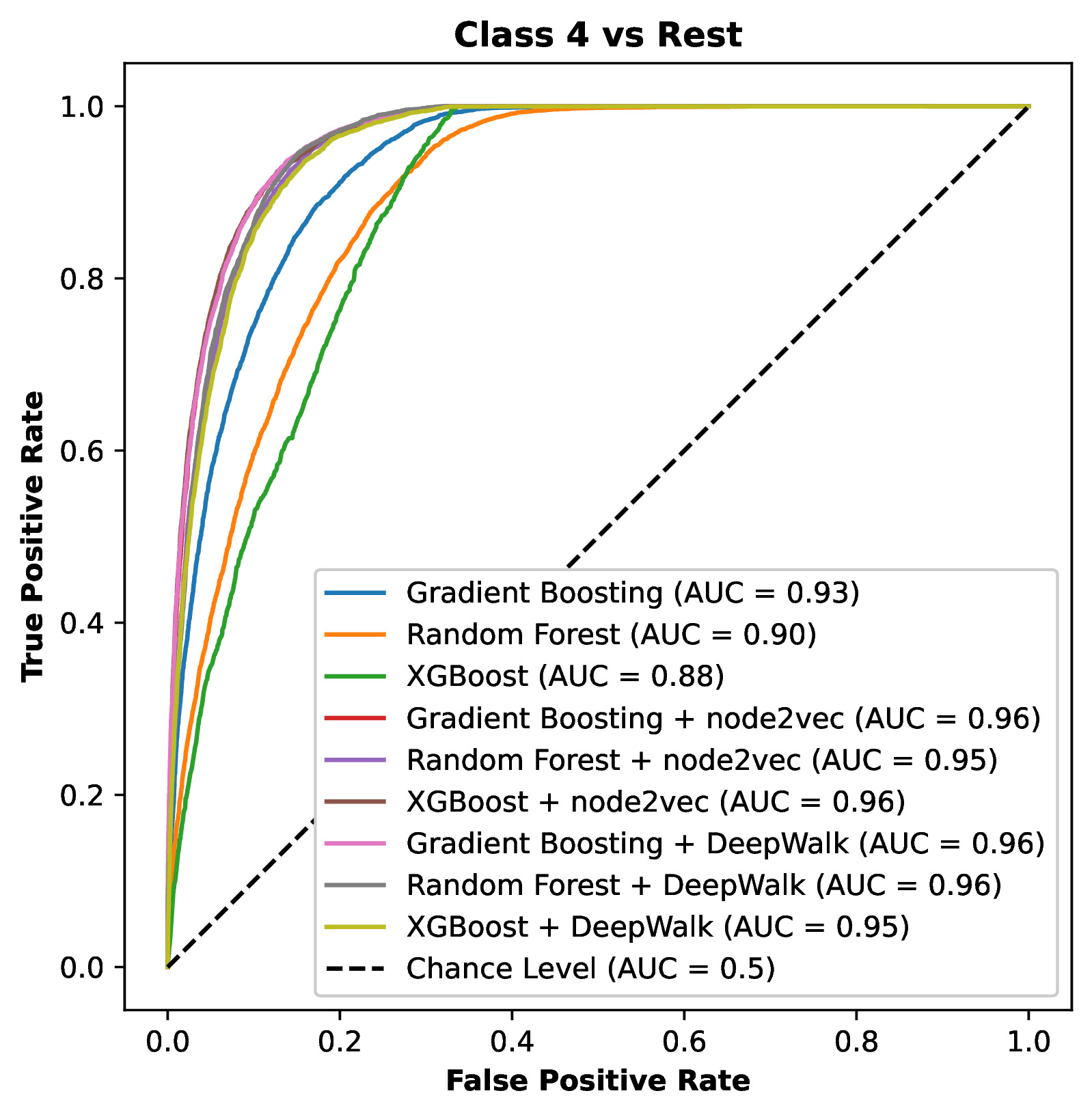}
\caption*{}
\end{minipage}
\vspace{-1cm}
\caption{One-vs-Rest ROC curves and AUC score for all models.}
\label{roc_auc}
\end{figure*}

Table~\ref{cl_rf_cr} presents the results of nine experiments, as we have three predictive models and three variations of structural features: no structural features, node2vec (+ degree and eigenvector), and DeepWalk (+ degree and eigenvector). From the results presented in Table~\ref{cl_rf_cr}, we can make the following observations:

\begin{itemize}
    \item [\ding{233}] Incorporating structural features using either DeepWalk or node2vec significantly improves the performance of all three prediction models (Random Forest, XGBoost, and Gradient Boosting) in terms of Accuracy, Precision, Recall, and F1-score.

      \item [\ding{233}]  Gradient Boosting consistently outperforms Random Forest and XGBoost across all variations, achieving the best overall performance when combined with node2vec structural features.

      \item [\ding{233}]  There is a close performance between the models incorporating DeepWalk and node2vec features, with only a slight edge in favor of the node2vec-based models.

        \item [\ding{233}] Given the large number of students enrolled in MOOCs, even a 3\% increase in performance is quite significant.

      \item [\ding{233}]  The use of structural features leads to a substantial improvement in the model performance. For instance, the Gradient Boosting model's performance increases from an F1-score of 0.89 to 0.92 when incorporating node2vec features.

 \end{itemize}   

These observations highlight the importance of considering complex structural relationships between entities in MOOC grade prediction models and the potential of graph-based representation learning techniques to enhance the performance of these models.

Figure~\ref{roc_auc} presents the ROC curves and their corresponding AUCs. Since we are addressing a multi-class classification problem, ROC curves can be plotted by comparing one class against the rest. As a result, for each class, we plotted one-versus-rest ROC curves for all nine models and calculated the AUC scores. It can be observed that the predictions for the middle classes are nearly perfect for all models. However, for classes 0 and 4, the models' performance is not as strong. Nevertheless, the ROC curves demonstrate that the models incorporating structural features outperform others in these classes.


\begin{table*}[h]
\setlength{\extrarowheight}{0.1pt}
\setlength\tabcolsep{3pt}
\centering
\caption{\label{gb_cr}Detailed classification results for different Gradient Boosting variations}
\begin{tabular}{l|ccc||ccc||ccc} 
&  \multicolumn{3}{|c||}{\textbf{Gradient Boosting}}&  \multicolumn{3}{|c||}{\textbf{Gradient Boosting + node2vec}}&  \multicolumn{3}{|c}{\textbf{Gradient Boosting + DeepWalk}} \\ \hline
 Class & Precision & Recall & F1-score& Precision & Recall & F1-score& Precision & Recall & F1-score \\\hline
class 0  &    0.68 &     0.72 &      0.70 & 0.82 &     0.72 &      \textbf{0.77}&    0.83 &     0.72 &      \textbf{0.77} \\
class 1   &   0.99  &    0.99 &     0.99  &0.99  &    1.00 &     \textbf{1.00}     &   0.99  &    1.00 &     \textbf{1.00}   \\
class 2   &  0.99   &   1.00  &    0.99   &  0.99   &   1.00  &    0.99 &  0.99   &   1.00  &    \textbf{1.00}    \\
class 3     & 0.99    &  1.00   &   \textbf{1.00 }& 0.99    &  1.00   &   \textbf{1.00}       & 0.99    &  1.00   &   \textbf{1.00}      \\
class 4     & 0.79     & 0.74    &  0.76   & 0.82     & 0.88    & \textbf{ 0.85} & 0.82     & 0.89    &  \textbf{0.85 }   \\ \hline
accuracy   & & &                      0.84   &&&  0.89 &&& 0.89\\
macro avg    &   0.89  &    0.89   &   0.89 &   0.92  &    0.92   &   0.92 &   0.92  &    0.92   &   0.92  \\
weighted avg    &   0.84    &  0.84  &    0.84   &   0.89    &  0.89  &    0.89    &   0.89    &  0.89  &    0.89     \\

\end{tabular}

\end{table*}
\begin{figure*}[htbp]
\centering
\noindent\begin{minipage}[b]{.33\textwidth}
\centering
\includegraphics[width=1\linewidth]{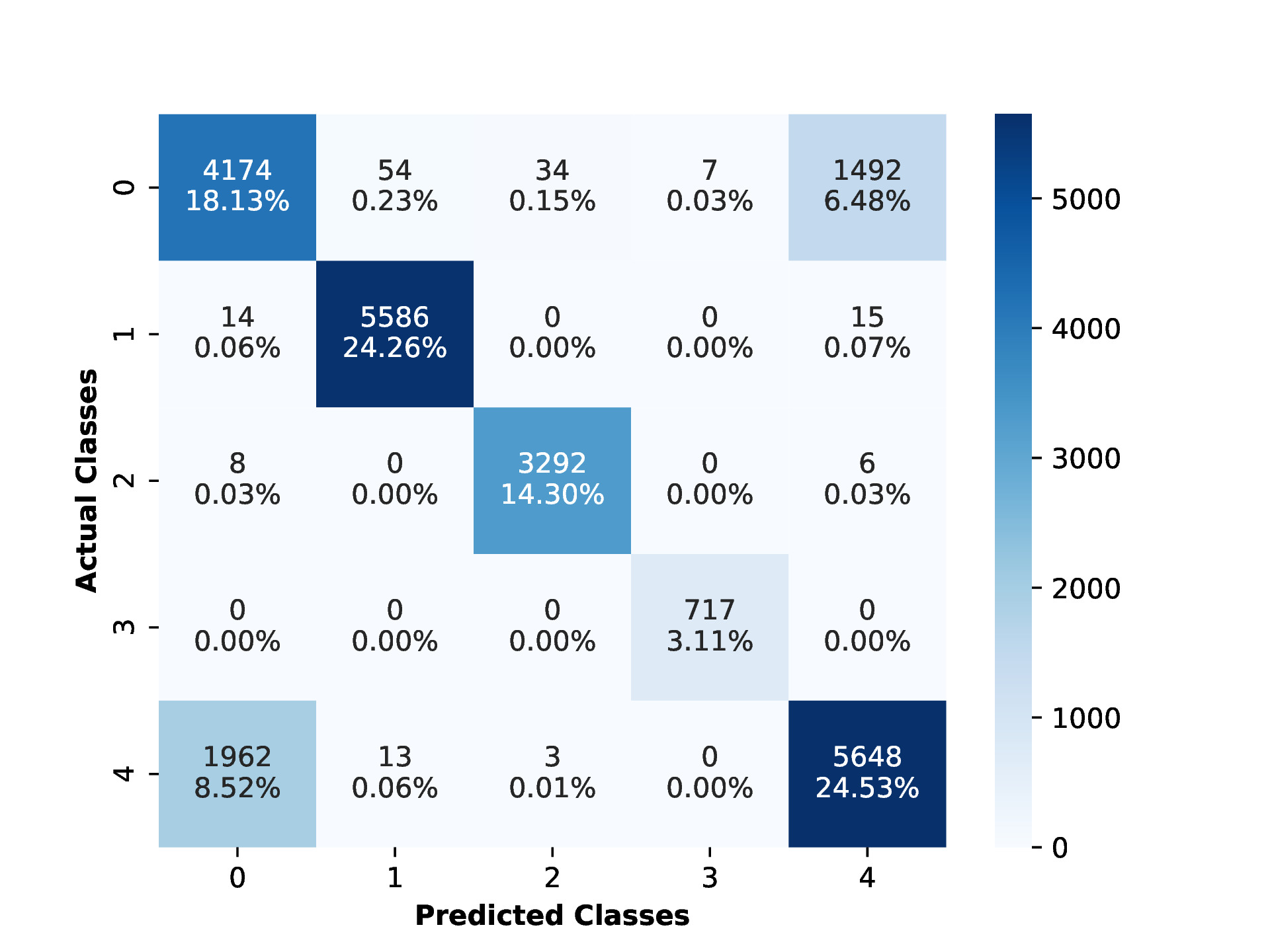}
\caption*{(a)}
\end{minipage}\hfill
\begin{minipage}[b]{.33\textwidth}
\centering
\includegraphics[width=1\linewidth]{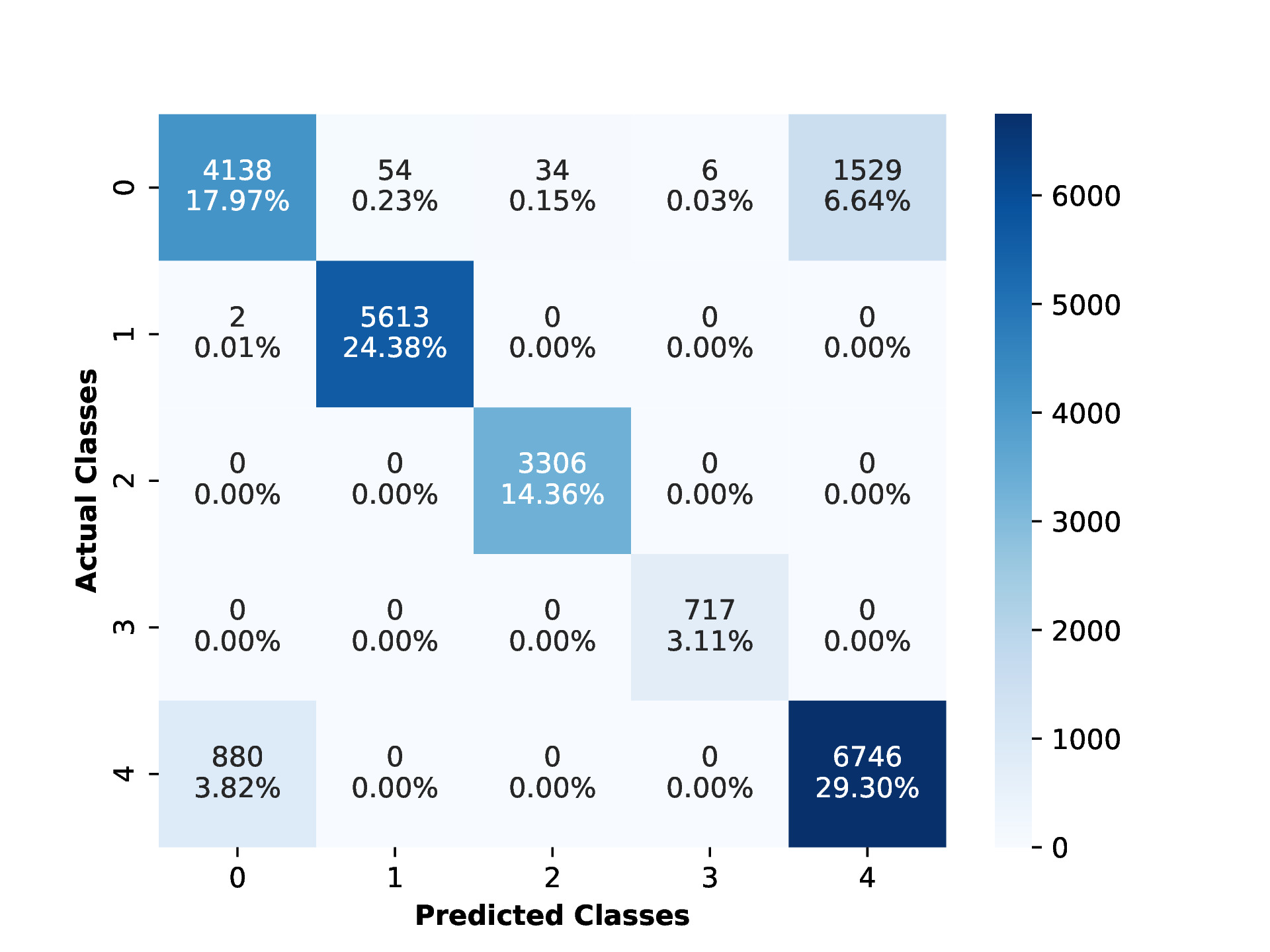}
\caption*{(b)}
\end{minipage}\hfill
\begin{minipage}[b]{.33\textwidth}
\centering
\includegraphics[width=1\linewidth]{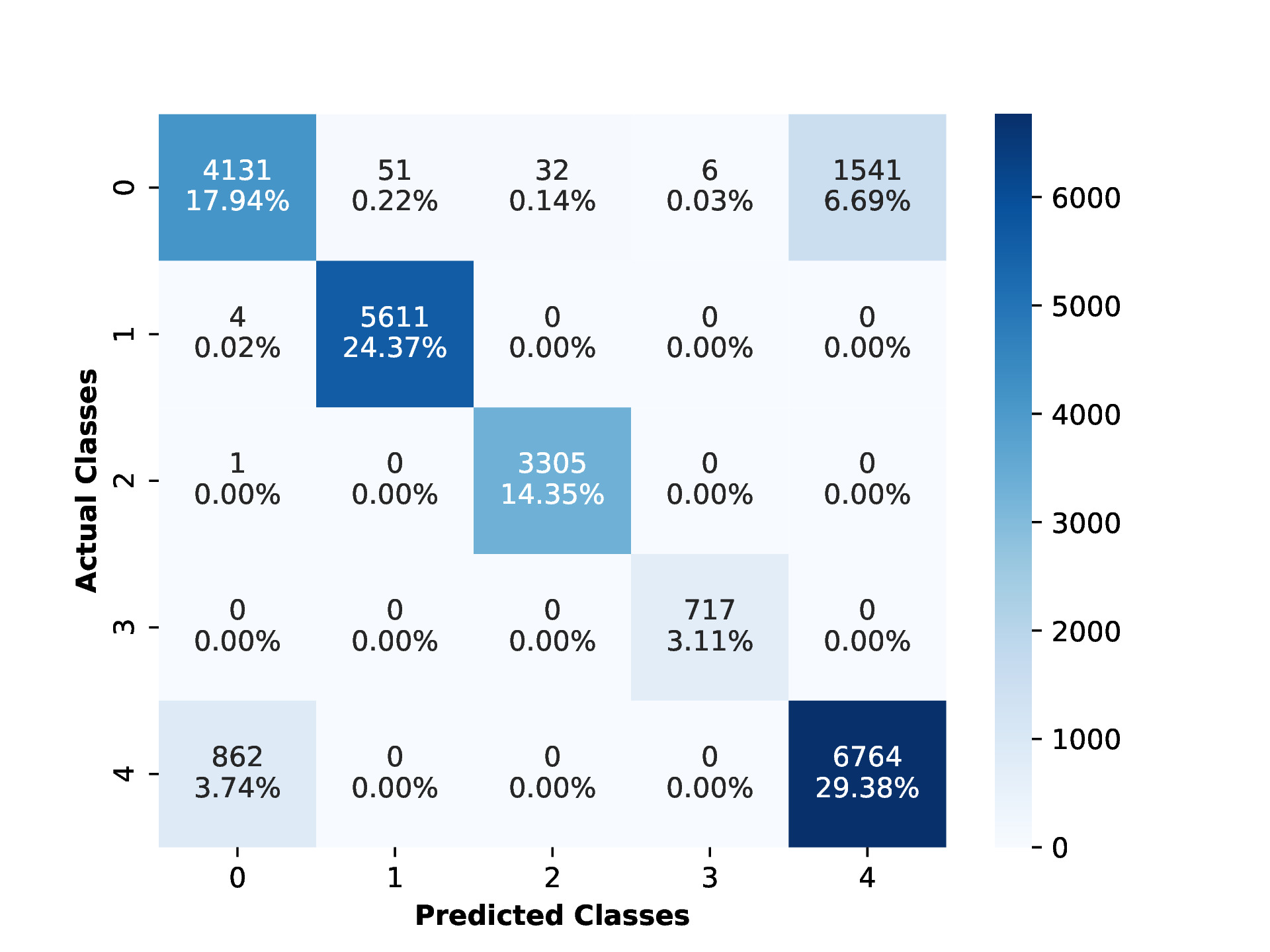}
\caption*{(c)}
\end{minipage}
\caption{Confusion matrix of (a) Gradient boosting model. (b) Gradient boosting + node2vec model. (c) Gradient boosting + DeepwWalk model.}
\label{confusion}
\end{figure*}
\begin{figure}[htbp]
\centering
\includegraphics[width=0.9\linewidth]{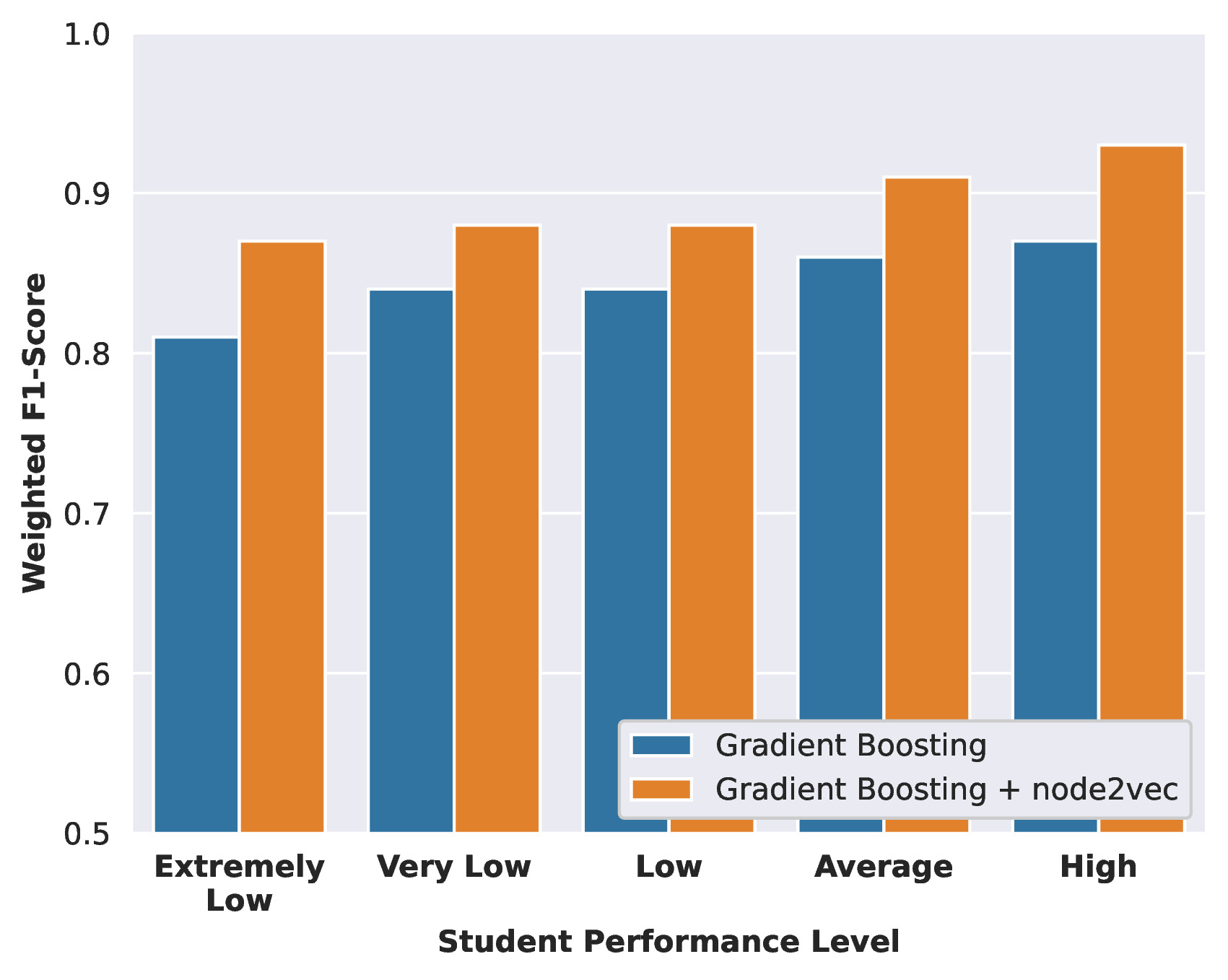}
\caption{Performance of Gradient Boosting and Gradient Boosting + node2vec models for different categories of students.}
\label{student_performace}
\end{figure}

Additionally, we have provided the classification reports (Table~\ref{gb_cr}) and the confusion matrices (Figure~\ref{confusion}) for Gradient Boosting (the top-performing model) with and without using structural features to examine their effectiveness on the \texttt{challenge} grade prediction task in greater detail. The primary classification metrics are displayed as a function of each class in the classification report. This offers a more comprehensive understanding of the classifier's behavior than global accuracy, which can conceal functional deficiencies in one class of a multi-class problem. We make the following observations from Table~\ref{gb_cr} and Figure~\ref{confusion}:

\begin{itemize}
  \item [\ding{233}] We can observe that structural features significantly improve the performance of classes 0 and 4, which the basic Gradient Boosting model struggles to predict effectively.

   \item [\ding{233}] While there has not been a substantial improvement in performance for classes 1, 2, and 3, it's important to note that incorporating structural features makes a notable difference in predicting classes 0 and 4.

    \item [\ding{233}] The basic Gradient Boosting model faces difficulties in accurately predicting these classes, but the inclusion of structural features has resulted in a 7\% improvement in performance for class 0 and a 9\% increase for class 4.

    \item [\ding{233}] It is crucial to emphasize the improvement in class 0, as this class represents students with the lowest grades, whose accurate prediction in automated grade prediction in MOOCs is essential~\cite{yang2017behavior,adnan2021predicting}. These students are considered at-risk, and identifying them accurately can enable targeted interventions and support to help them succeed in their courses. The incorporation of structural features in the prediction models proves to be a valuable approach to enhance the prediction performance for this critical group of students. The following section offers an in-depth explanation of our automated \texttt{challenge} grade prediction system across  student groups with different academic attainment levels. 
\end{itemize}


\subsection{Grade prediction and student academic performance }
\label{subsec:student_levels}
As discussed in Section \ref{sec:introduction}, the primary goal of grade prediction is typically to identify and support students with low academic performance. It is essential for a grade prediction model to accurately predict future grades for struggling students. To evaluate the performance of our best model (Gradient Boosting) for low-performing students, we categorized students into five groups based on their grades in the training set, which can be considered as their previous performance. Students with more than 90\% of their grades in classes 0 and 1 (low grades) were labeled as ``Extremely Low" performing. Similarly, students with 80-90\% low grades were labeled ``Very Low", 50-80\% low grades as ``Low", 20-50\% low grades as ``Average", and students with less than 20\% low grades were labeled ``High" performing.

Figure~\ref{student_performace} illustrates the performance of Gradient Boosting with and without using structural features for the five different categories of students. In terms of Weighted F1-score, the prediction for ``Extremely Low" performing and ``Very Low" performing students improved by 6\% and 4\%, respectively, when structural features were included. This, again, demonstrates the positive impact of incorporating structural features for predicting the performance of struggling students.

\subsection{Feature Analysis}
\label{subsec:features}
\vspace{-0.1cm}

\begin{figure*}[htbp]
\centering
\noindent\begin{minipage}[b]{.33\textwidth}
\centering
\includegraphics[width=1\linewidth]{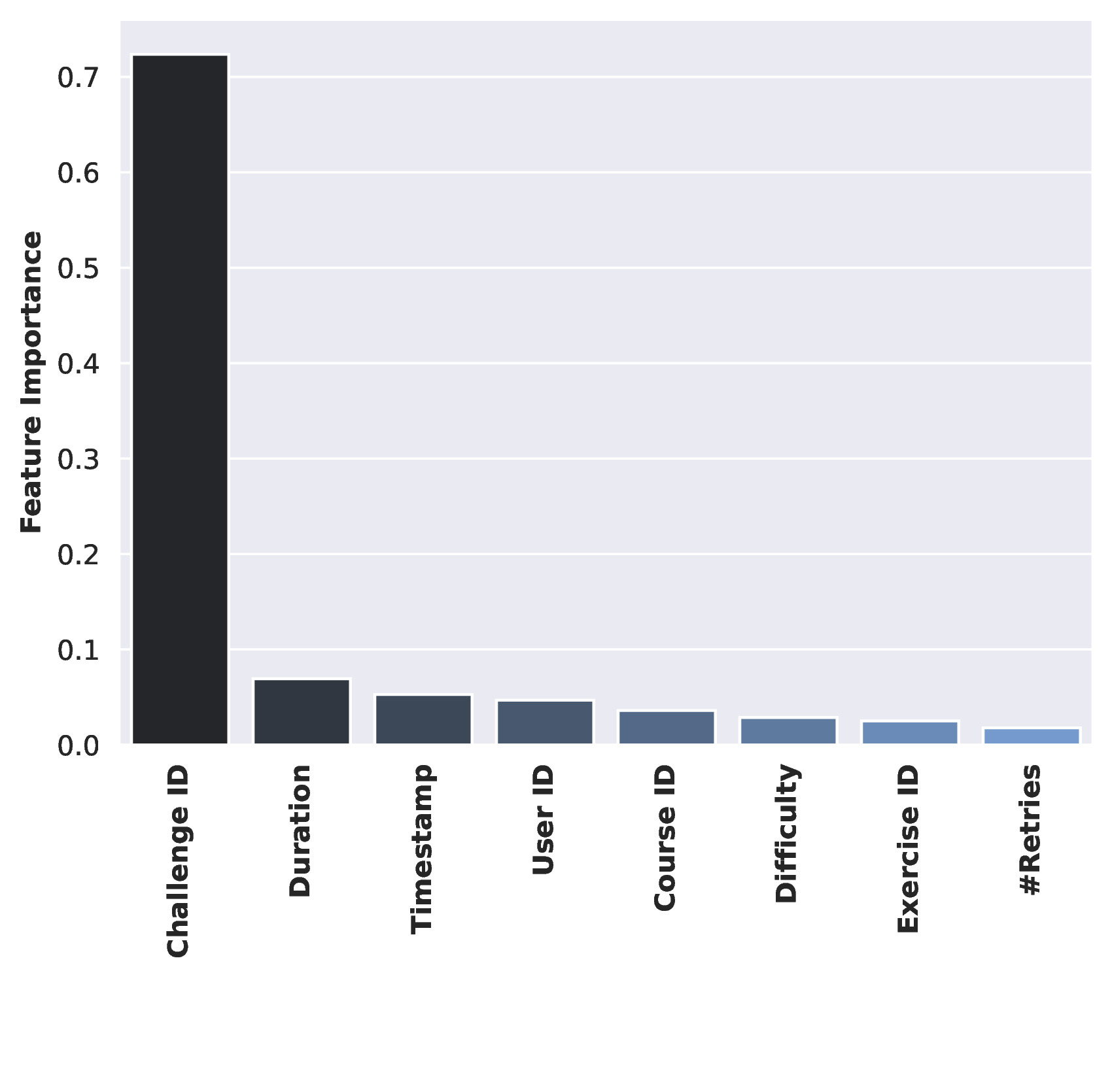}
\caption*{(a)}
\end{minipage}\hfill
\begin{minipage}[b]{.33\textwidth}
\centering
\includegraphics[width=1\linewidth]{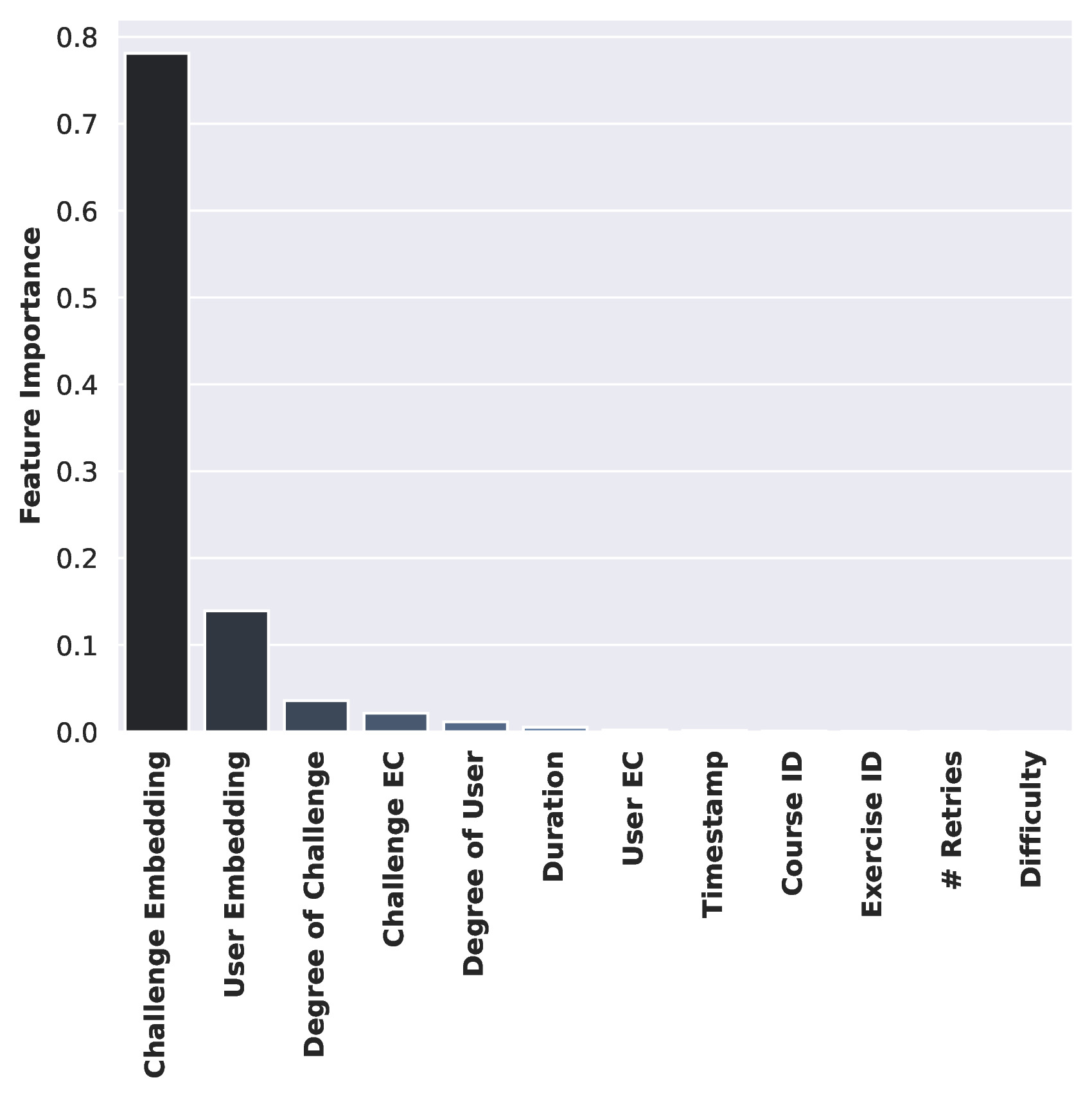}
\caption*{(b)}
\end{minipage}\hfill
\begin{minipage}[b]{.33\textwidth}
\centering
\includegraphics[width=1\linewidth]{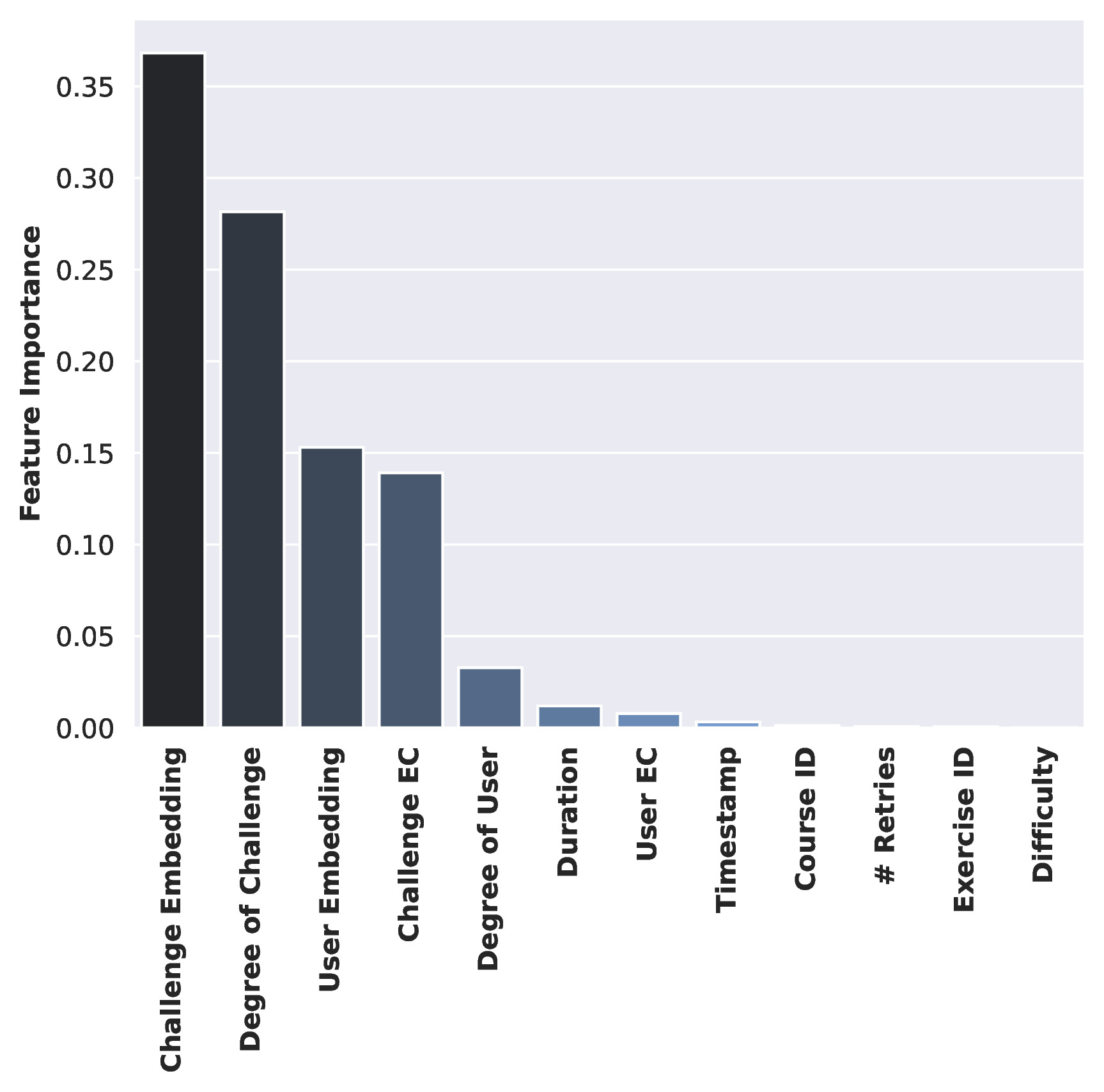}
\caption*{(c)}
\end{minipage}

\caption{ Feature importance of models (a) Gradient boosting model. (b) Gradient boosting + node2vec model. (c) Gradient boosting + DeepWalk model.}
\label{feature_importance}
\end{figure*}

Figure~\ref{feature_importance} shows the feature importance in different Gradient Boosting models. In the first model, \texttt{challenge} ID, duration, and timestamp are the top three factors in predicting the \texttt{challenge} grade. However, in the second model, the embedding of \texttt{challenge} nodes along with user node embeddings and degree of \texttt{challenge} node dominates the feature importance plot. In other words, structural features become prominent features that surpass ordinary MOOC-related features, e.g., \texttt{challenge} ID. Finally, in the third model, where DeepWalk was used to obtain the embeddings, importance is more distributed but still the \texttt{challenge} embeddings, and user embeddings along with node degree and eigenvector centrality are critical factors in the model's predictions. This, again, emphasizes the importance of considering structural features in grade prediction.
\section{Conclusion}
\label{sec:conclusion}

MOOC is one of the most exciting recent phenomena in higher education. The promise of MOOC is the democratization of education~\cite{dillahunt2014democratizing}. It has attracted millions of learners worldwide. The high enrollment, diversity of topics, and the scarcity or lack of instructors and student-instructor interaction have automated some tasks. In particular, grade prediction is a crucial task providing several advantages. However, in this study, we identified two issues associated with grade prediction in MOOC: 1) the lack of enough granularity in the prediction (i.e., focusing on high-level tasks such as course prediction), and 2) ignoring complex yet effective structural relationships among pertinent entities. In this study, we first introduced a novel and large MOOC dataset providing multiple aspects of academic data to overcome these issues. Notably, \texttt{challenges} are short and small exercises focusing on a specific skill. Since this is a complicated problem and offers high granularity in the prediction, we pressed onward to predict students' \texttt{challenge} grades. We extracted salient and dense structural representations from a constructed interaction graph between students and \texttt{challenges} to address the second issue and improve prediction performance. We performed extensive experiments and showed that structural features could significantly improve the quality of the \texttt{challenge} prediction.

There are several intriguing future directions. First, an interesting research question is how our \texttt{challenge} prediction system is perceived in real-world settings. In fact, our research group is currently investigating this aspect. Second, one could extend the interaction graph beyond a bipartite graph and include other entities, such as courses and disciplines. Third, such a heterogeneous graph could be used to perform other tasks, like course recommendation. Finally, exploring more advanced machine learning methods, such as node representation learning in heterogeneous graphs~\cite{wang2022survey}, applied to the graph in our dataset, is a promising direction for future research.
\section*{Acknowledgment}
In this work, Dr. Jiangtao Huang is supported by the National Natural Science Foundation of China (62067007).
 \balance
 \bibliographystyle{IEEEtran}
 \begin{spacing}{0.90}
  \bibliography{references}
  \end{spacing}
\end{document}